%% file: main.tex
\documentclass[conference]{IEEEtran}
\IEEEoverridecommandlockouts
\usepackage{cite}
\usepackage{amsmath,amssymb,amsfonts}
\usepackage{algorithm,algorithmic}
\usepackage{graphicx}
\usepackage{textcomp}
\usepackage{xcolor}
\usepackage{multirow}
\usepackage{comment}
\usepackage{subcaption}
\usepackage{diagbox}
\usepackage[draft]{hyperref}
\hypersetup{draft}
\usepackage{url}
\def\BibTeX{{\rm B\kern-.05em{\sc i\kern-.025em b}\kern-.08em
    T\kern-.1667em\lower.7ex\hbox{E}\kern-.125emX}}
\begin{document}

\title{Efficient Compressed Ratio Estimation Using Online Sequential Learning for Edge Computing\\
%{\footnotesize \textsuperscript{*}Note: Sub-titles are not captured in Xplore and
%should not be used}
%\thanks{Identify applicable funding agency here. If none, delete this.}
}

\IEEEoverridecommandlockouts\IEEEpubid{\makebox[\columnwidth]{978-1-6654-6483-3/23~\copyright~2023 IEEE \hfill} \hspace{\columnsep}\makebox[\columnwidth]{ }}

%\begin{comment}
\author{\IEEEauthorblockN{Hiroki Oikawa, Hangli Ge, Noboru Koshizuka}
\IEEEauthorblockA{
Graduate School of Interdisciplinary Information Studies,\\
The University of Tokyo,\\
Tokyo, Japan\\
\{hiroki.oikawa, hangli.ge, noboru\}@koshizuka-lab.org}}
%\end{comment}

\maketitle

\begin{abstract}
Owing to the widespread adoption of the Internet of Things, a vast amount of sensor information is being acquired in real time.
Accordingly, the communication cost of data from edge devices is increasing.
Compressed sensing (\textit{CS}), a data compression method that can be used on edge devices, has been attracting attention as a method to reduce communication costs.
In CS, estimating the appropriate compression ratio is important.
There is a method to adaptively estimate the compression ratio for the acquired data using reinforcement learning (\textit{RL}).
However, the computational costs associated with existing RL methods that can be utilized on edges are often high. 
In this study, we developed an efficient RL method for edge devices, referred to as the actor--critic online sequential extreme learning machine (\textit{AC-OSELM}), and a system to compress data by estimating an appropriate compression ratio on the edge using AC-OSELM.
The performance of the proposed method in estimating the compression ratio is evaluated by comparing it with other RL methods for edge devices.
The experimental results indicate that AC-OSELM demonstrated the same or better compression performance and faster compression ratio estimation than the existing methods.
\end{abstract}

\begin{IEEEkeywords}
edge computing, data compression, IoT
\end{IEEEkeywords}

\newcommand{\argmin}{\mathop{\rm arg~min}\limits}
\newcommand{\argmax}{\mathop{\rm arg~max}\limits}
\newcommand{\minimize}{\mathop{\rm minimize}\limits}

\section{Introduction}
The widespread use of Internet of Things ({\textit{IoT}}) devices has led to the acquisition of diverse sensor information in real time.
Utilizing such sensor information, technologies such as artificial intelligence have made significant progress.
However, with the increase in the number of IoT devices, the volume of data communication is also increasing, resulting in immense power and communication bandwidth costs\cite{background}.
\par
Data compression technology is used to reduce the cost of data transmission by reducing the amount of data without compromising its features.
In particular, compressed sensing (\textit{CS}) has garnered attention as a method that can be used on edge devices with low computational resources\cite{compressed}.
In CS, it is important to determine an appropriate compression ratio for the acquired data considering the trade-off between the reduction of data transmission and reconstruction errors.
\par
A method exists for estimating the compression ratio in CS using reinforcement learning (\textit{RL}).
Sekine et al. proposed a method for estimating the optimal compression ratio using deep RL (\textit{DRL}) for edge devices\cite{lacsle}.
DRL is an RL method that uses deep neural networks (\textit{DNN}) and has made significant progress in recent years\cite{dqn}.
However, edge devices with fewer computational resources, such as sensor nodes, have difficulty incorporating DRL, which is computationally expensive.
Therefore, a more efficient RL method is required to estimate the compression ratio on edge devices such as sensor nodes.
\par
There exist efficient RL methods for edge devices that employ the online sequential extreme learning machine ({\textit{OS-ELM}}) \cite{oselm}.
The OS-ELM is a single-layer neural network capable of fast sequential learning.
Watanabe et al. proposed the OS-ELM-Q-Network ({\textit{OS-QNet}}) as an RL method with low computational cost, based on OS-ELM\cite{oselm-q-net}.
They demonstrated that OS-QNet could be efficiently executed on field-programmable gate arrays (\textit{FPGAs}).
However, in OS-QNet, the cost of inferring the optimal action becomes large when there are many action patterns, such as the compression ratio.
Therefore, the estimation of compression ratios using OS-QNet at the edge becomes a computational bottleneck in data compression using CS.
\par
This study proposes an efficient RL method referred to as the actor--critic OS-ELM ({\textit{AC-OSELM}}).
An actor--critic is an RL method that sequentially trains an actor to learn a policy and a critic to estimate the value of a state\cite{actor-critic}.
In the proposed AC-OSELM, the actor and critic, which are both composed of single-layer neural nets, are updated with a deterministic policy gradient (\textit{DPG}) \cite{dpg} and OS-ELM, respectively, to achieve RL with a low computational cost.
AC-OSELM can efficiently infer optimal actions even when the number of action patterns is large, such as the compression ratio.
Therefore, we propose a system that uses AC-OSELM to estimate the compression ratio in CS.
This system allows efficient data compression on edge devices with low computational resources.\par
The contributions of this study are summarized as follows:
\begin{itemize}
    \item We developed AC-OSELM as an RL method for edge devices. Unlike the existing OS-QNet, AC-OSELM can be used efficiently when there are several action patterns.
    \item We developed AC-OSELM-based compression ratio estimation system for CS.
    This system has a low computational cost and is suitable for execution on edge devices.
    \item We conducted experiments to evaluate the performance of the compression ratio estimation in CS using AC-OSELM and compared it with OS-QNet.
\end{itemize}

The experimental results indicated that AC-OSELM achieved the same or better compression performance than OS-QNet for every dataset.
In addition, AC-OSELM significantly reduced the computational cost of inference compared with OS-QNet, despite a higher computational cost of learning.
The compression ratio estimation model for edge devices requires higher real-time performance for inference than for learning.
Therefore, the proposed compression ratio estimation method based on AC-OSELM can be considered suitable for edge-device implementation.

\section{Related Work}
Various studies have been conducted on data compression on edge devices.
Azar et al. proposed an efficient compression scheme using Squeeze, which is a fast error-bounded lossy compression scheme\cite{sz}, for IoT devices\cite{compression_iot}.
CS is a method that can be applied to sparse data\cite{compressed}.
This compression method achieves accurate data recovery and low compression ratios for sparse data.
Owing to its low computational cost, CS has been used in applications wherein real-time performance is required, such as image communication systems for IoT monitoring applications at sensor nodes\cite{compressed_sensing_monitoring}.
Li et al. also proposed a power- and communication-efficient framework for wireless sensor networks and IoT devices using CS\cite{compressed_signal}.
CS requires an appropriate compression ratio for the data to perform both data reduction and reconstruction properly.
A system that uses a DNN on edge devices to estimate the compression ratio has been proposed previously\cite{lacsle}.
Using RL, this system can learn and estimate an appropriate compression ratio for the acquired data.
However, owing to the high computational costs associated with DNNs, the model cannot be used on edge devices with low computational resources.
Therefore, efficient RL models are required for use at the network edge. \par
As mentioned, OS-ELM is a neural network with a low computational cost that follows changes in data trends and has a fast sequential learning capability for data acquired in a sequence\cite{oselm}.
It can achieve high-speed processing on FPGAs, making it suitable for use on edge devices\cite{fpga}.
OS-QNet is an RL method that uses the OS-ELM\cite{oselm-q-net}.
This method is known to be computationally efficient and significantly faster than the existing Deep Q-Network (\textit{DQN}) \cite{dqn} when implemented on FPGAs.
However, the computational cost of selecting actions in OS-QNet increases in proportion to the number of actions, and it cannot handle continuous action spaces, such as the compression ratio.
Therefore, this study proposes AC-OSELM as an OS-ELM-based RL method that can be used even when the pattern of actions is not finite.
In addition, we propose AC-OSELM-based system for estimating the compression ratio in CS that can be applied to edge devices with small computational resources. 
\par
Data extraction is a method for reducing the data volume at the edge.
This method reduces the data sent to the cloud server by selecting data from edge devices.
Papageorgiou et al. proposed a system for efficiently extracting acquired time series data by automatically switching between multiple data handlers\cite{nec}.
In addition, a method exists for extracting the representative data from a large data stream by using submodular optimization\cite{submodular}.
Such methods that directly reduce the number of data to be transmitted can be used in conjunction with data compression methods that can reduce the cost of transmission without reducing the quantity of data. \par

\section{Preliminaries}
\subsection{Compressed Sensing}
CS is a data compression method proposed by Donoho\cite{compressed}.
This method compresses the acquired $n$-dimensional data $x\in\mathbb{R}^{n\times1}$ into $m$-dimensional data $y\in\mathbb{R}^{m\times1}$ using a random matrix $\Phi\in\mathbb{R}^{m\times n}$, as shown in the following equation:

\begin{align}
  y = \Phi x
\end{align}

Because $m < n$, determining the value of $x$ from the values of $y$ and $\Phi$ is usually impossible.
We assume that $x$ is sparse in a certain feature space.
Then, using the transformation matrix $\Psi \in \mathbb{R}^{n\times n}$ and sparse vector $x_s \in \mathbb{R}^{n\times1}$, $x$ can be expressed as follows:

\begin{align}
  x = \Psi x_s
\end{align}

The value of $x_s$ can be obtained by solving the L1-norm minimization problem for $x_s$, as follows:

%\vspace{-5mm}
\begin{align}
  %\minimize{||x_s||_{1}} ~ {\rm subject~to~} y = \Phi\Psi x_s
  &\minimize{||x_s||_{1}} \\
  &{\rm subject~to~} y = \Phi\Psi x_s
\end{align}

Using $x_s$, we can recover the value $x$ of the acquired data as $x = \Psi x_s$. 
Thus, CS can achieve fast compression for sparse data in a certain feature space.
This is a suitable method for implementation on edge devices wherein fast processing is required to maintain real-time performance. \par

\subsection{Reinforcement Learning}
In RL, an agent in an environment observes its current state and learns an action strategy (policy) that maximizes the reward obtained from the environment for the observed state.

\subsection{OS-QNet}
Q-learning is an RL method estimating the effectiveness of an action $a_t$ in a certain state $s_t$ using the Q-function $Q(s_t,a_t)$.
In Q-learning, $Q(s_t,a_t)$ is sequentially updated, and for state $s_t$, the action $a_t$ that maximizes $Q(s_t,a_t)$ is selected.
DQN is a method for constructing this Q-function using DNN\cite{dqn}.
Although DQN demonstrates superior generalization performance, it incurs a high computational cost and is not suitable for execution on edge devices.
Therefore, OS-QNet, a Q-learning method using the OS-ELM, has been proposed\cite{oselm-q-net}.

\subsubsection{OS-ELM}
~\par
\textbf{Algorithm of ELM}: 
The ELM is a single-layered neural network model that can be trained analytically\cite{elm}.
It has the following parameters: $\alpha$, the weight from the input layer to the hidden layer; $b$, the bias of the hidden layer; and $\beta$, the weight from the hidden layer to the output layer.
Only $\beta$ is updated during the learning process, while the other values are fixed.
In the training process, the value of $\beta$ that minimizes the following value $L$ for input data $X$ and target data $Y$ is calculated: 

%\vspace{-5mm}
\begin{align}
    L = ||H\beta - Y|| \\
    H = g(X\alpha+b)
\end{align}

\noindent
where $g(\cdot)$ denotes the activation function of the hidden layer. \par
$\hat{\beta}$, which minimizes $L$, can be obtained analytically using the following equation:

%\vspace{-5mm}
\begin{align}
    \hat{\beta} = H^{\dag}Y
\end{align}

\noindent
where $H^{\dag}$ is the Moore--Penrose pseudo inverse of $H$.
A method to train the Q-function and the critic function (to be described later) with ELM has been proposed and is known to be efficient for learning\cite{elm-q-net,elm-critic}.

\textbf{Algorithm of OS-ELM}: 
The OS-ELM is a model that extends ELM to sequential learning\cite{oselm}.
Let the sequentially acquired data be denoted as $(X_i,Y_i) (i=0,1,...)$, $\hat{\beta}_{i}$ where the weights from the hidden layer to the output layer in the OS-ELM are updated, as shown in the following equation:

%\vspace{-4mm}
\begin{align}
P_0 &= \left(H_0^T H_0 \right)^{-1},~\hat{\beta}_{0} = P_0H_0^TY_0 \\
P_{i+1} &= P_i - P_i H_{i+1}^T \left(\lambda I + H_{i+1} P_i H_{i+1}^T \right)^{-1}H_{i+1} P_i \\
 \hat{\beta}_{i+1} &=   \hat{\beta}_{i} + P_{i+1}H_{i+1}^T \left(Y_{i+1} - H_{i+1}  \hat{\beta}_{i} \right)
\end{align}

\noindent
where $\lambda$ denotes the forgetting rate, which is a constant to be set in advance.

\subsubsection{Algorithm of OS-QNet}

%\begin{comment}
\begin{figure}[b]
\begin{minipage}[t]{0.999\hsize}
\centering
\includegraphics[scale=0.15]{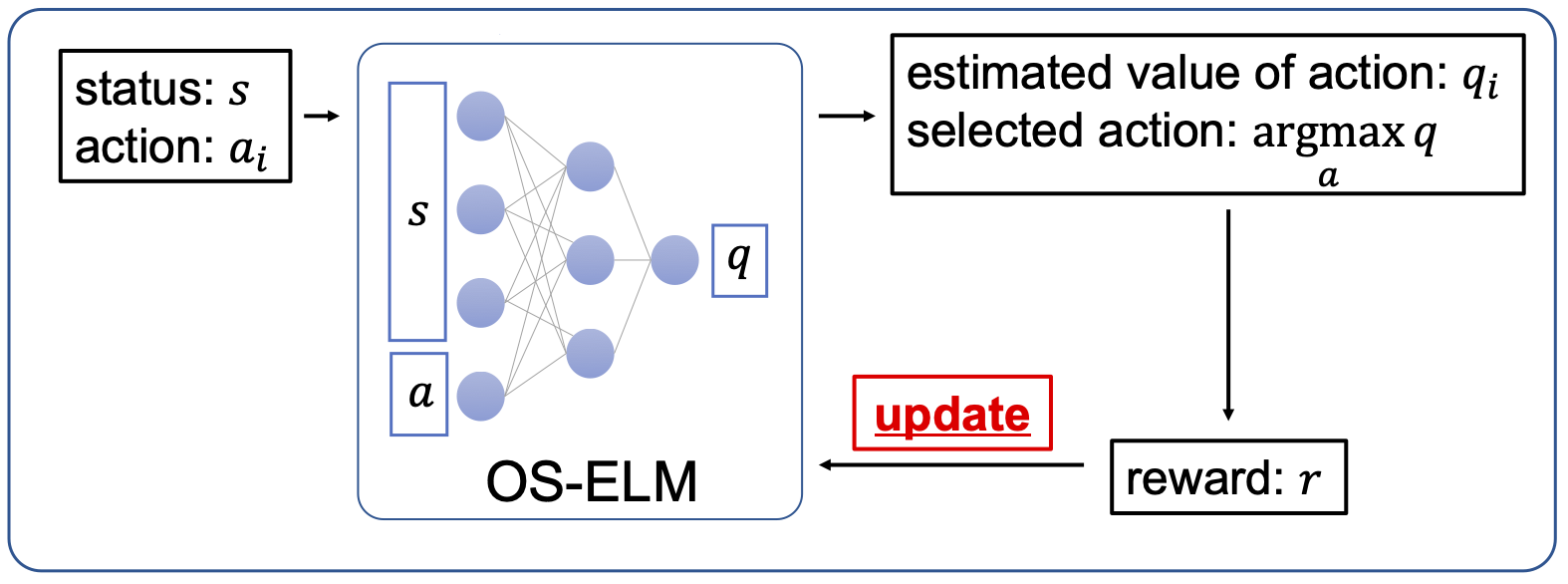}
\end{minipage}
\caption{Overview of OS-ELM-Q-Network}
\label{fig:os-qnet}
%\vspace{-5mm}
\end{figure}
%\end{comment}

Figure~\ref{fig:os-qnet} depicts an overview of OS-QNet.
In OS-QNet, the Q-function is constructed using OS-ELM.
This algorithm is shown in Algorithm~\ref{alg:os-elm-q-net}
where $\alpha$, $b$, $\beta$, and $N$ correspond to the weight from the input layer to the hidden layer, bias in the hidden layer, weight from the hidden layer to the output layer, and the update period, respectively.
$\epsilon(t)$ is the probability of choosing a random action and decays as $t$ increases.
The update of $\beta$ in Algorithm~\ref{alg:os-elm-q-net} is given by the following equation.
When $t = N$, it is updated according to Eq.(11); otherwise, it is updated according to Eq.(12,13).

%\vspace{-4mm}
\begin{align}
P &= \left(H^T H \right)^{-1},~\beta = PH^TR \\
P &= P - PH^T \left(\lambda I + HPH^T \right)^{-1}HP \\
\beta &= \beta + PH^T \left(R - H\beta \right)
\end{align}

\noindent
where $H$ and $R$ are computed using Eq.(14) using $(a_t, s_t,r_t)$ in $D$.

%\vspace{-4mm}
\begin{align}
H = \left(
    \begin{array}{ccc}
      h|_{s=s_1,a=a_1} \\
      \vdots \\
      h|_{s=s_N,a=a_N}
    \end{array}
  \right)
,
R = \left(
    \begin{array}{ccc}
      r_1 \\
      \vdots \\
      r_N
    \end{array}
  \right)
\end{align}

\noindent
where $h|_{s=s_t,a=a_t}$ is the output of the hidden layer of OS-ELM for $(s_t, r_t)$ and corresponds to $g((s_t, r_t)\alpha+b)$ using the activation function $g(\cdot)$.\par

\begin{algorithm}[b]
 \small
 \caption{Algorithm of OS-QNet}
  \label{alg:os-elm-q-net}
 \begin{algorithmic}[0]
 %\renewcommand{\algorithmicrequire}{\textbf{Input:}}
 %\renewcommand{\algorithmicensure}{\textbf{Output:}}
 %\REQUIRE m, $T[1,2,...,n]$, $X[1,2,...,n]$
 %\ENSURE  $T_{p}[1,2,...,m]$, $X_{p}[1,2...,m]$
 \STATE Initialize parameters $\theta = \{\alpha,\beta,b\}$ using random values $\mathbb{R} \in [0,1]$, buffer $D = \{\}$, global step $t = 0$
 %\STATE Initialize buffer $D = \{\}$
 %\STATE Initialize global step $t = 0$
 \FOR{$episode$ $\in$ \{1,2,...\}}
   \STATE Observe($s_t$)
   \WHILE{$s_t$ $\neq$ end}
     \STATE $t \leftarrow t+1$, $p$ $\sim$ $U(0,1)$
     \IF{$p$ $<$ $\epsilon$}
       \STATE $a_t \leftarrow$ random action value
     \ELSE
       \STATE $a_t = \argmax_{a\in A}{Q_{\theta}{(s_t,a)}}$
     \ENDIF
   \ENDWHILE
   \STATE Observe($s_{t+1}$,$r_t$) from environment
   \STATE $r_t$ $\leftarrow$ $r_t + \gamma \max_{a\in A}{Q_{\theta}{(s_{t+1},a)}}$
   \STATE Store($a_{t}$,$s_{t}$,$s_{t+1}$,$r_t$) in $D$
   \IF{$t \% N = 0$}
     \STATE Update $\beta$ using $D$ according to Eq.(11, 12, 13)
     \STATE $D = \{\}$
   \ENDIF
 \ENDFOR
\end{algorithmic} 
\end{algorithm}

At a low computational cost, OS-QNet can learn based on this procedure and be implemented on edge devices.
However, when estimating the optimal action for a state, all action patterns must be evaluated with a Q-function.
Therefore, OS-QNet cannot handle continuous action spaces, such as compression ratio, which is a continuous value.
In addition, even if there are a finite number of action patterns, when there are several, the computational cost when inferring the action becomes large.

\section{Actor--critic online sequential extreme learning machine (AC-OSELM)}
In this study, we developed AC-OSELM, a novel RL method for edge devices, and a system to compress data by estimating an appropriate compression ratio on the edge using AC-OSELM.

\begin{figure}[b]
\begin{minipage}[t]{0.999\hsize}
\centering
\includegraphics[scale=0.13]{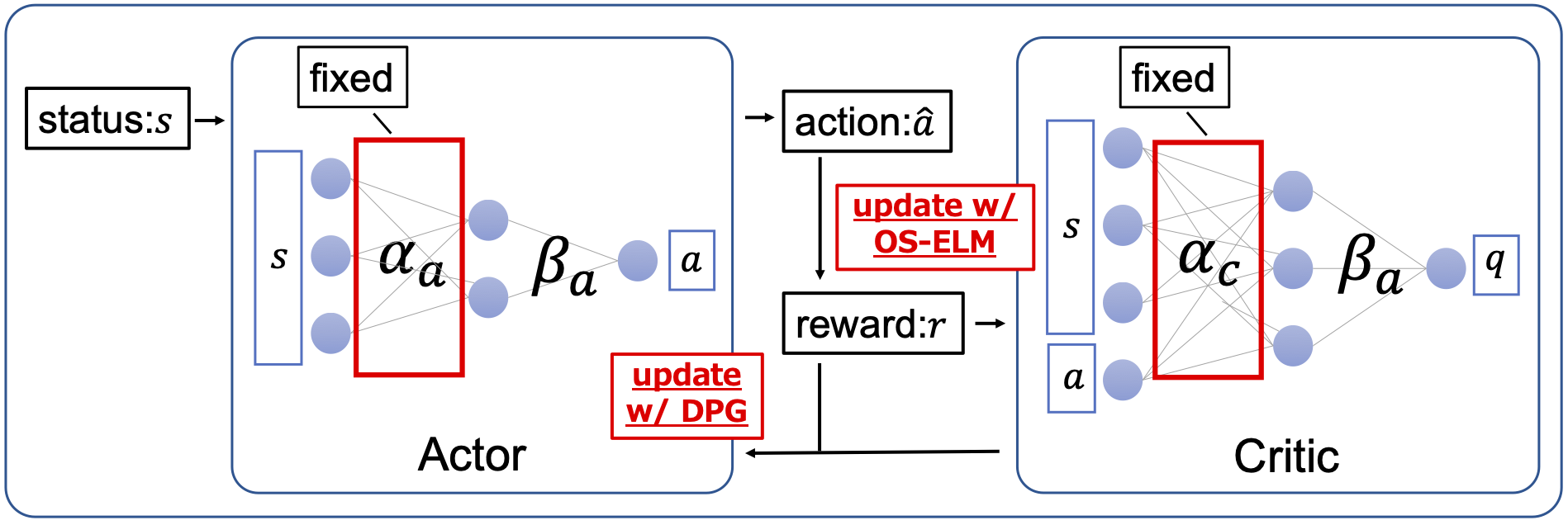}
%\vspace{-4mm}
\end{minipage}
\caption{Overview of Actor-Critic-OS-ELM}
\label{fig:ac-oselm}
%\vspace{-5mm}
\end{figure}

\subsection{Algorithm of the proposed AC-OSELM}
The actor--critic is a method that simultaneously learns the actor $\mu(s_t)$ that determines the action, and the critic $Q(s_t,a_t)$ that evaluates the effectiveness of the policy\cite{actor-critic}.
In general, actor--critic methods are known to make the model robust to noise and stabilize learning. \par
In the proposed AC-OSELM, the actor and critic are both consist of a single-layer neural network, as shown in Figure ~\ref{fig:ac-oselm}.
For both the actor and critic, the weights from the input layer to the hidden layer are fixed, and only the weights from the hidden layer to the output layer are updated.
The critic and actor are updated using OS-ELM and DPG\cite{dpg}, respectively. \par
Let the dimensionality of the state, the dimensionality of the action, the size of the hidden layer, the activation function in the hidden layer, and the activation function in the output layer be $D$, $k$, $m_a$, $g_{ah}(\cdot)$, and $g_{ao}(\cdot)$, respectively.
Using $s \in \mathbb{R}^{1\times D}$, $\alpha_a \in \mathbb{R}^{D\times m_a}$, $b_a \in \mathbb{R}^{1\times m_a}$, and $\beta_a \in \mathbb{R}^{m_a\times k}$, the function of the actor $\mu_{\theta_a}(s)$ is as follows:

%\vspace{-5mm}
\begin{align}
    h_a = g_{ah}(s\alpha_a+b_a) \in &\mathbb{R}^{1\times m_a}\\
    \mu_{\theta_a}(s) = g_{ao}(h_a\beta_a) \in &\mathbb{R}^{1\times k}
\end{align}

\noindent
where $\alpha_a$, $b_a$, and $\beta_a$ correspond to the weights from the input to the hidden layer of the actor, bias in the hidden layer, and weights from the hidden to the output layer, respectively. \par
Let the size of the hidden layer and activation function in the hidden layer be $m_c$ and $g_{ch}(\cdot)$, respectively.
Using $a \in \mathbb{R}^{1\times k}$, $\alpha_c \in \mathbb{R}^{(D+k)\times m_c}$, $b_c \in \mathbb{R}^{1\times m_c}$, and $\beta_c \in \mathbb{R}^{m_c\times 1}$, the function of critic $Q_{\theta_c}(s,a)$ is as follows:

%\vspace{-5mm}
\begin{align}
    x = (s~a) &\in \mathbb{R}^{1\times (D+k)}\\
    h_c = g_{ch}(x\alpha_c+b_c) &\in \mathbb{R}^{1\times m_c}\\
    Q_{\theta_c}(s,a) = h_c\beta_c &\in \mathbb{R}^{1\times 1}
\end{align}

\noindent
where $\alpha_c$, $b_c$, and $\beta_c$ correspond to the weights from the input layer to the hidden layer of the critic, the bias in the hidden layer, and weights from the hidden layer to the output layer, respectively. \par
The algorithm for AC-OSELM implementation is described in Algorithm~\ref{alg:ac-oselm}.
Here, $N$ denotes the update period.
$\epsilon$ is the noise added to the value of the action output from the actor and decays as $t$ increases. \par

\begin{algorithm}[b]
 \small
 \caption{Algorithm of AC-OSELM}
  \label{alg:ac-oselm}
 \begin{algorithmic}[0]
 %\renewcommand{\algorithmicrequire}{\textbf{Input:}}
 %\renewcommand{\algorithmicensure}{\textbf{Output:}}
 %\REQUIRE m, $T[1,2,...,n]$, $X[1,2,...,n]$
 %\ENSURE  $T_{p}[1,2,...,m]$, $X_{p}[1,2...,m]$
 \STATE Initialize parameters $\theta_a = \{\alpha_a,b_a,\beta_a\}$, $\theta_c = \{\alpha_c,b_c,\beta_c\}$ using random values $\mathbb{R} \in [0,1]$, buffer $D = \{\}$, global step $t = 0$
 %\STATE Initialize buffer $D = \{\}$
 %\STATE Initialize global step $t = 0$
 \FOR{$episode$ $\in$ \{1,2,...\}}
   \STATE Observe($s_t$)
   \WHILE{$s_t$ $\neq$ end}
     \STATE $t \leftarrow t+1$
     \STATE $a_t = \mu_{\theta_a}(s_t) + \epsilon ~(\epsilon \sim P(\epsilon|t))$
   \STATE Observe($s_{t+1}$,$r_t$) from environment
   \STATE $r_t$ $\leftarrow$ $r_t + \gamma Q_{\theta_c}{(s_{t+1},\mu_{\theta_a}(s_{t+1}))}$
   \STATE Store($a_{t}$,$s_{t}$,$s_{t+1}$,$r_t$) in $D$
   \IF{$t \% n = 0$}
     \STATE Update $\beta_c$ using $D$ according to Eq. (11--13)
     \STATE Update $\beta_a$ using $D$ according to Eq. (22--25)
     \STATE $D = \{\}$
   \ENDIF
   \ENDWHILE
 \ENDFOR
\end{algorithmic} 
\end{algorithm}

$\beta_c$ and $\beta_a$ in Algorithm~\ref{alg:ac-oselm} are updated as follows.
First, $\beta_c$ is updated according to the OS-ELM using Eq. (11--13).
This is the same as the update of the Q-function of OS-QNet.

$\beta_a$ is updated according to DPG.
In DPG, the parameter $\theta_c$ of $Q$ is fixed and the parameter $\theta_a$ of $\mu$ changes with the gradient of increasing $Q(s,\mu(s))$\cite{dpg}.
This gradient is given by the following equation:

%\vspace{-4mm}
\begin{align}
    &\frac{\partial Q(s,a|\theta_c)}{\partial \theta_a} = \frac{\partial a}{\partial \theta_a}\cdot \frac{\partial Q(s,a|\theta_c)}{\partial a} \\
    \approx \frac{1}{N}&\sum_{i=1}^{N} \frac{\partial \mu(s|\theta_a)}{\partial \theta_a}|_{s=s_i}\cdot \frac{\partial Q(s,a|\theta_c)}{\partial a}|_{s=s_i,a=\mu(s_i|\theta_a)}
\end{align}

In AC-OSELM, the weights from the input layer to the hidden layer $\alpha_a$ and the bias in the hidden layer $b_a$ are fixed, and the weight from the hidden layer to the output layer $\beta_a$ is updated according to DPG.
The formula for updating $\beta_a$ is as follows.
Here, $\eta$ is the update rate.

%\vspace{-4mm}
\begin{align}
    \frac{\partial y}{\partial \beta_a} &\approx \frac{1}{N}\sum_{i=1}^N \frac{\partial y}{\partial \beta_a}|_{s = s_i} = \mathrm{d}\beta_a\\
    \beta_a &= \beta_a + \eta \mathrm{d}\beta_a
    %\beta_a &= \beta_a + \eta (\mathrm{d}\beta_a - \tau\beta_a)
\end{align}

\noindent
$\dfrac{\partial y}{\partial \beta_a}$ is calculated by the following equation, where $\otimes$ corresponds to the Hadamard product.

%\vspace{-4mm}
\begin{align}
    %\frac{\partial y}{\partial \beta_a} &= h_a^T g'_{ao}(h_a\beta_a) \otimes (g'_{ch}(s\alpha_c+b_c)\otimes \beta_c^T)\alpha_{ck}^T \\
    \frac{\partial y}{\partial \beta_a} &= h_a^T g'_{ao}(h_a\beta_a) \otimes (g'_{ch}(s\alpha_c+b_c) (\beta_c\otimes\alpha_{ck}^T)) \\
    \alpha_{ck} &= \left(
    \begin{array}{ccc}
      {\alpha_c}_{(D+1)1} & \ldots & {\alpha_c}_{(D+1)m_c} \\
      \vdots & \ddots & \vdots \\
      {\alpha_c}_{(D+k)1} & \ldots & {\alpha_c}_{(D+k)m_c}
    \end{array}
  \right)
\end{align}

By updating only $\beta_a$ in this manner, the computational cost associated with the actor update is small.
In particular, as $k$ is $1$ in the compression ratio estimation, $g'_{ao}(h_a\beta_a)$ and $g'_{ch}(s\alpha_c+b_c) (\beta_c\otimes\alpha_{ck}^T)$ becomes scalars, and $\mathrm{d}\beta_a$ can be easily computed.

Thus, AC-OSELM can perform RL while incurring a low computational cost, which is suitable for use on edge devices.
In addition, unlike OS-QNet, AC-OSELM can handle the continuous action spaces and achieve fast action selection even with a large number of action patterns.

\subsection{Application}
\begin{figure}[b]
\begin{tabular}{cc}
\begin{minipage}[t]{0.23\hsize}
%\centering
\begin{center}
\includegraphics[scale=0.2]{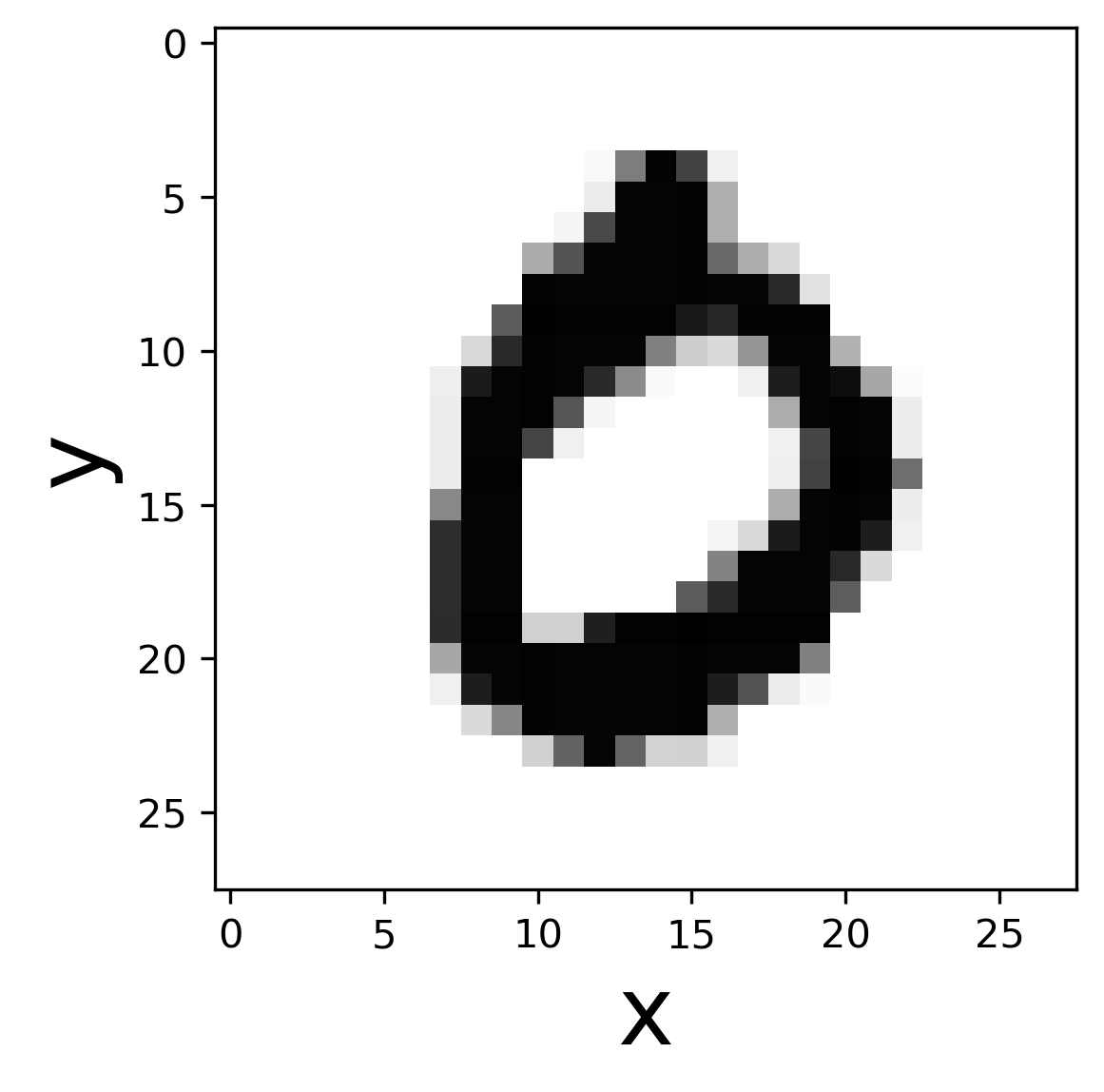}
\subcaption{the original image}
\end{center}
\end{minipage}
\begin{minipage}[t]{0.23\hsize}
%\centering
\begin{center}
\includegraphics[scale=0.2]{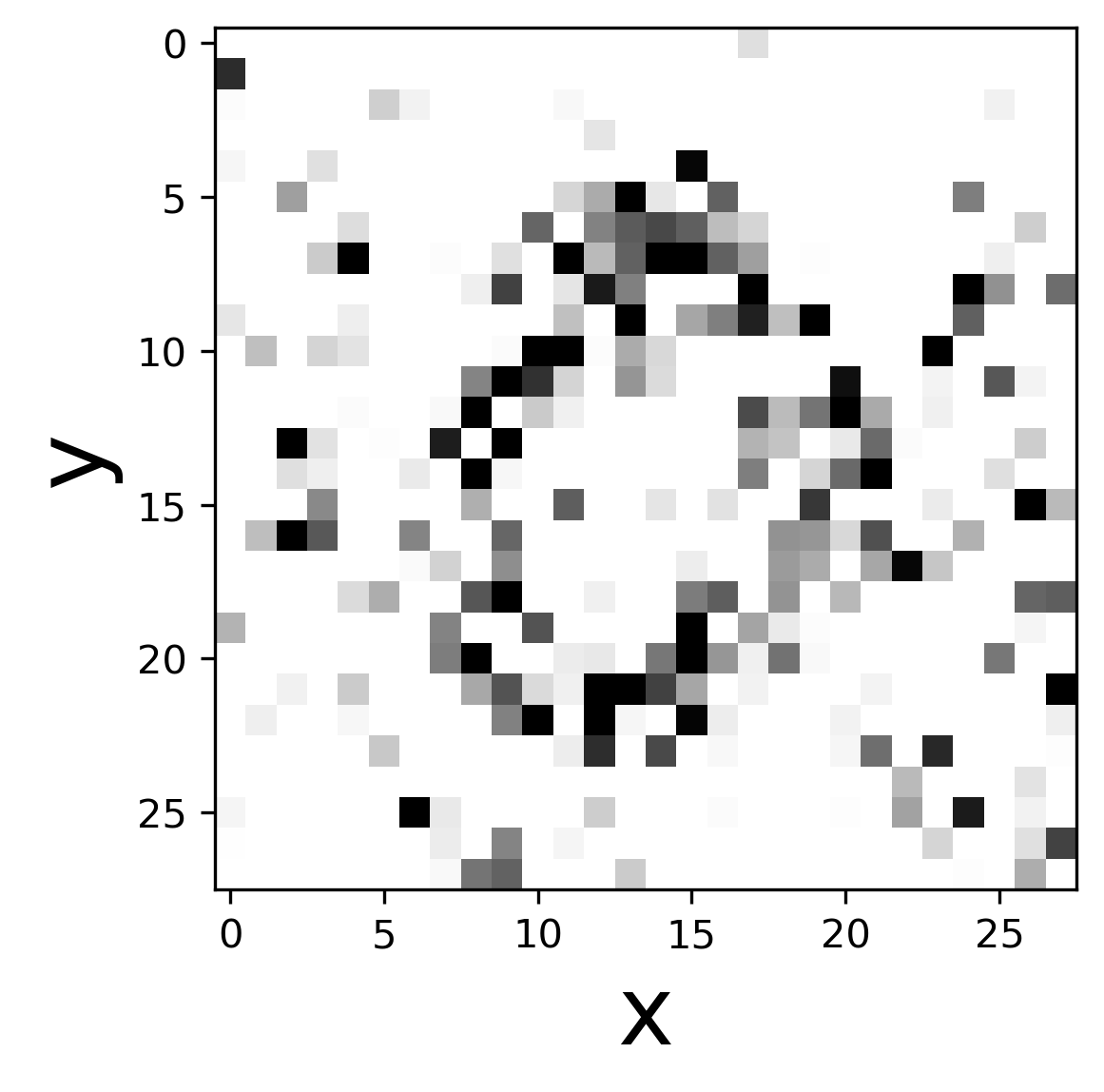}
\subcaption{$c$ = 0.25, $E(c)$ = 0.493}
\end{center}
\end{minipage}
\begin{minipage}[t]{0.23\hsize}
%\centering
\begin{center}
\includegraphics[scale=0.2]{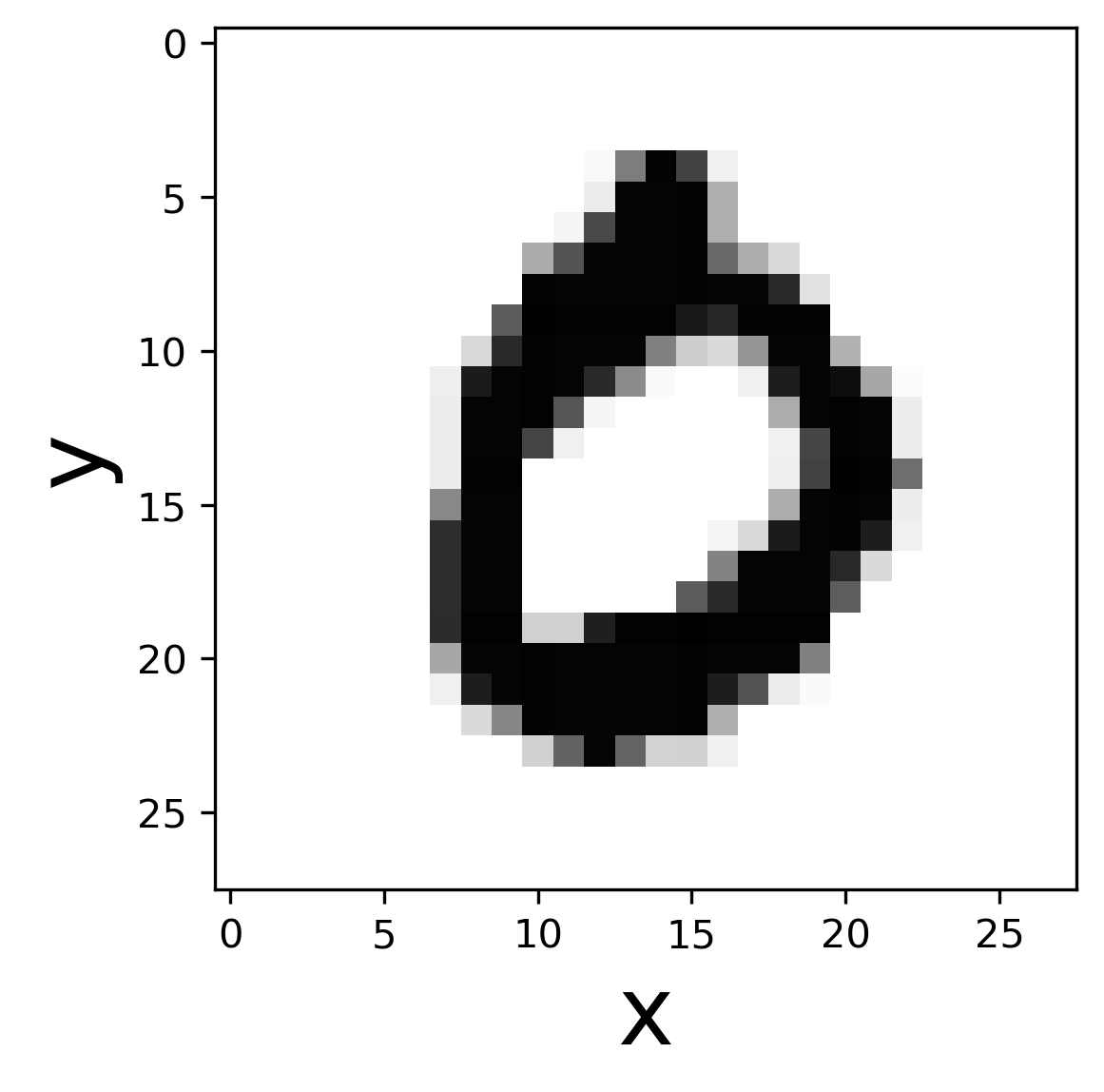}
\subcaption{$c$ = 0.50, $E(c)$ = \textbf{0.875}}
\end{center}
\end{minipage}
\begin{minipage}[t]{0.23\hsize}
%\centering
\begin{center}
\includegraphics[scale=0.2]{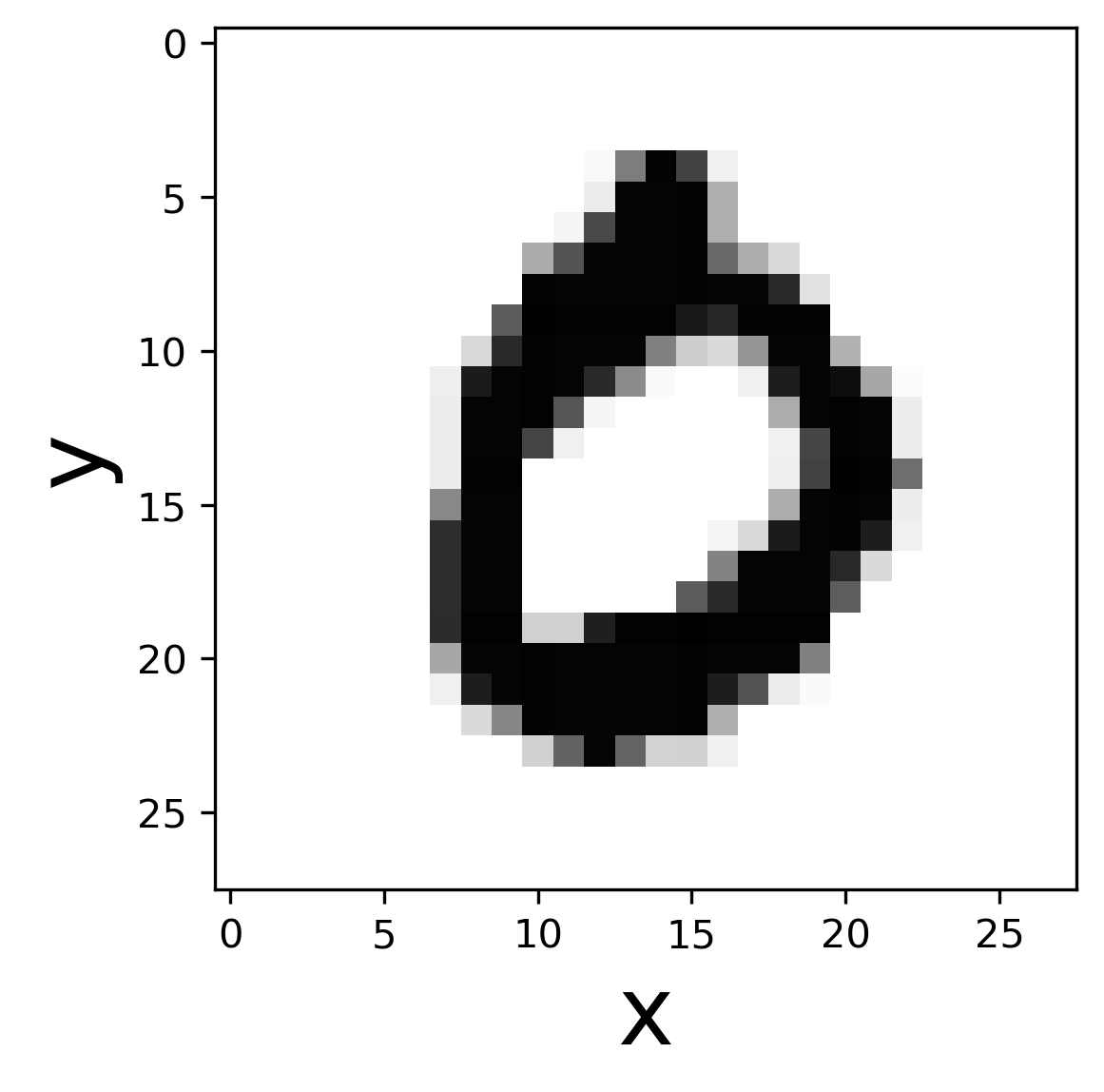}
\subcaption{$c$ = 0.75, $E(c)$ = 0.578}
\end{center}
\end{minipage}
%\caption{Reinforcement Learning}
\end{tabular}
\caption{The original image and images recovered from the compressed data at each compression rate}
\label{fig:mnist}
%\vspace{-5mm}
\end{figure}

This paper proposes using AC-OSELM to estimate the compression ratio in CS.
This allows for efficient data compression at the edges.
As stated, in CS, determining the appropriate compression ratio is necessary.
In particular, when processing on the edge, efficient communication can be achieved by adaptively changing the compression ratio according to the data being acquired.
However, a trade-off exists between the compression ratio and the data reconstruction error.
The smaller the compression ratio, the smaller the data communication cost; however, the data reconstruction error would increase.
Considering this trade-off, when estimating the compression ratio $c$ of CS by RL, the following function $E(c)$ was proposed as the data transmission efficiency\cite{lacsle}.

%\vspace{-5mm}
\begin{align*}
    E(c) &= k_1(-k_2 c^{k_3}+k_4-k_5 e(c)^{k_6}) \\
    e(c) &= max(0,RMSE(x, dec(x_c))-e_{thr})
\end{align*}

\noindent
where $x$, $x_c$, and $dec(x_c)$ corresponds to the original data, the data compressed $x$ with compression ratio $c$, and the data recovered from $x_c$, respectively.
$k_i (i=1,2,... ,6)$ and $e_{thr}$ are constant values that contribute to whether the reduction in the amount of data or the reduction of the reconstruction error is more important, and it is set according to the data and the application.
Figure~\ref{fig:mnist} depicts the original image in the MNIST dataset\cite{mnist} and images recovered from the compressed data at each compression ratio when $[k_1,k_2,k_3,k_4,k_5,k_6,e_{thr}] = [1,1,3,1,1.5,1,0.01]$.
The value of $k_i$ was determined such that the range of $E(c)$ was set to $[0,1]$ by referring to the value proposed in a prior work\cite{lacsle}.
The score is maximum when the compression ratio is $0.50$, which is a smaller value among the compression ratios that can properly restore the original data.
These results show that $E(c)$ functions properly as an evaluation index for the compression performance.\par

\begin{figure}[tb]
\begin{minipage}[t]{0.999\hsize}
\centering
\includegraphics[scale=0.12]{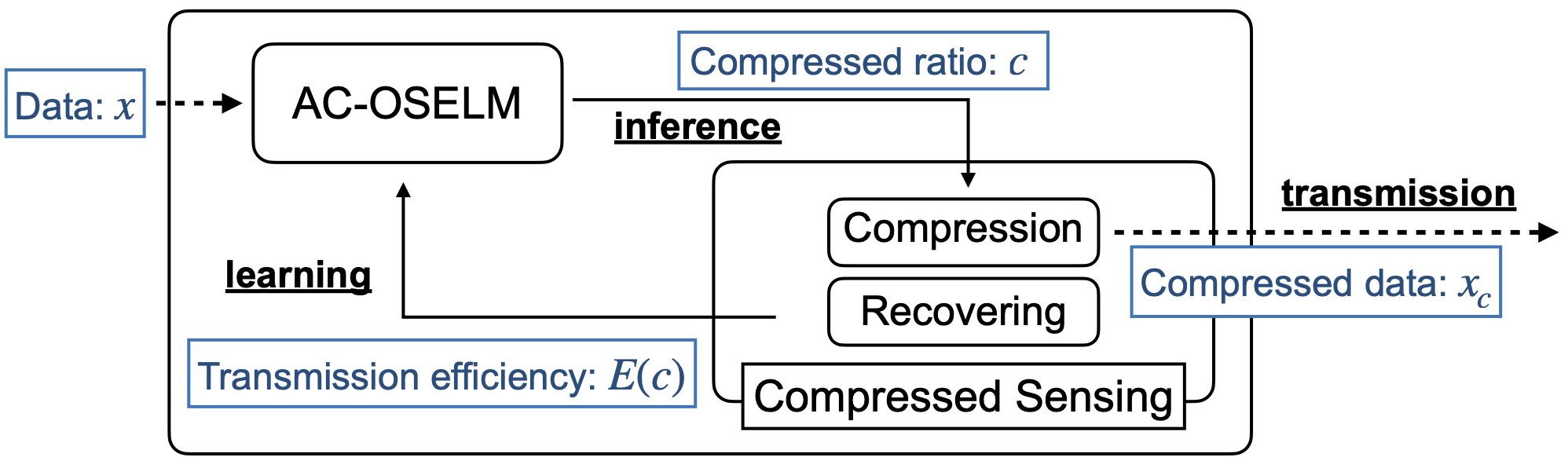}
\end{minipage}
\caption{Overview of the AC-OSELM-based compressed ratio estimation system}
\label{fig:ac-oselm-system}
%\vspace{-5mm}
\end{figure}

Figure~\ref{fig:ac-oselm-system} shows the overview of the proposed AC-OSELM-based compressed ratio estimation.
By training AC-OSELM with the acquired data as the input state, the compression ratio as the action, and $E(c)$ as the reward, the model to estimate the appropriate compression ratio according to the acquired data can be realized.

\section{Evaluation}

\subsection{Experimental Setup}

\textbf{Procedure}:
We evaluated the performance of AC-OSELM in estimating the compression ratio in CS compared with that of OS-QNet.
Each model acquires and compresses data sequentially in the experiment and is trained after every 10 data acquisitions
The acquisition of these 10 pieces of data is considered one step, and the compression accuracy at each step of each model is evaluated for 2,000 steps.
Each RL method was trained using $E(c)$ $([k_1,k_2,k_3,k_4,k_5,k_6,e_{thr}] = [1,1,3,1,1.5,1,0.01])$ as the reward.
We also compared the computational cost of each method to measure the computational time required to estimate the compression ratio and learn the model.
In AC-OSELM, the range of the compression ratio $c$ is set to $[0,1]$, and in OS-QNet, it is set to $\{0.1,0.2,...,1.0\}$.
The number of nodes in the middle layer and the forgetting rate of all models used in the experiment were set to 400 and 0.999, respectively. 
In data compression, the random matrix $\Phi$ was generated using $\mathcal{N}(0,1)$ and was fixed.

\begin{figure}[bt]
\begin{minipage}[t]{0.999\hsize}
%\centering
\begin{center}
\includegraphics[scale=0.6]{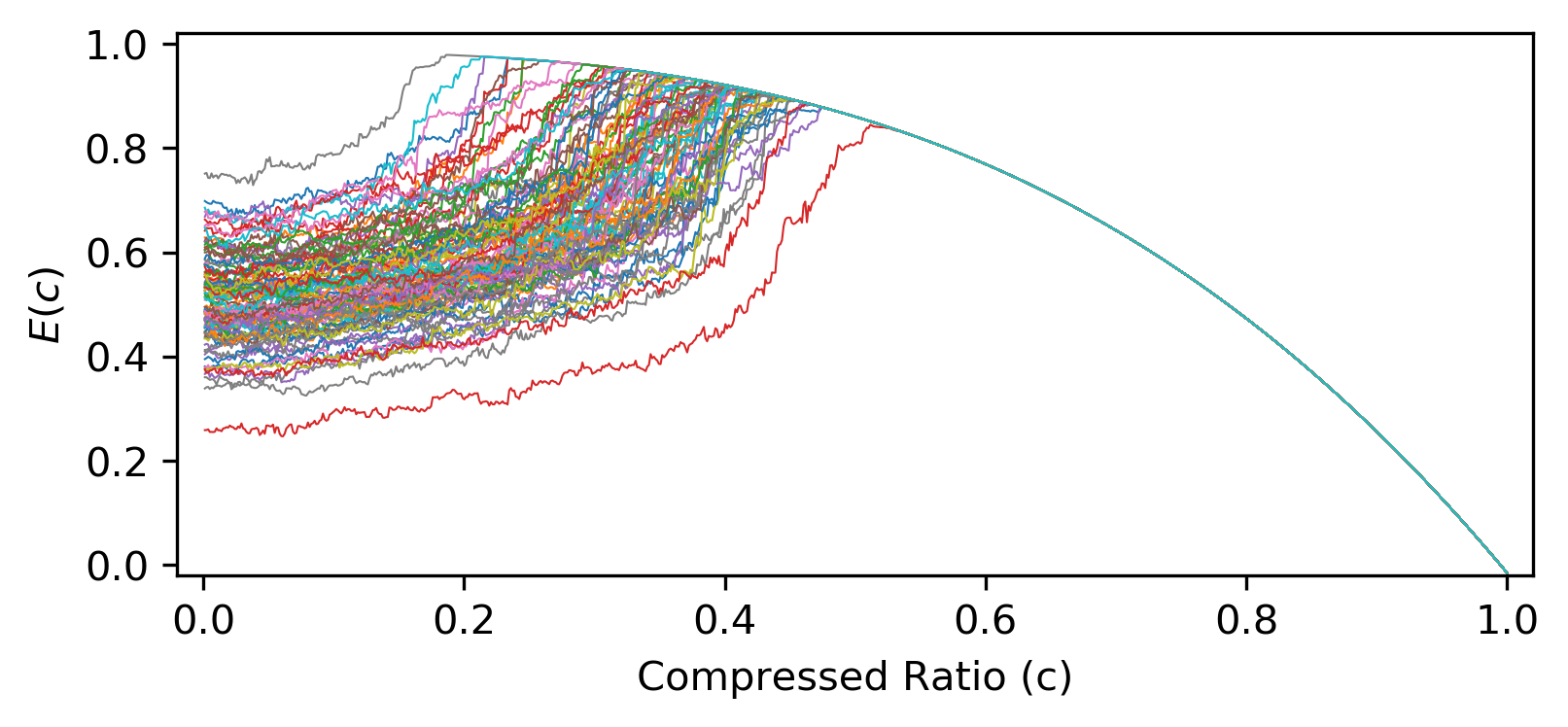}
\subcaption{MNIST}
\end{center}
\end{minipage}
\begin{minipage}[t]{0.999\hsize}
\centering
\begin{center}
\includegraphics[scale=0.6]{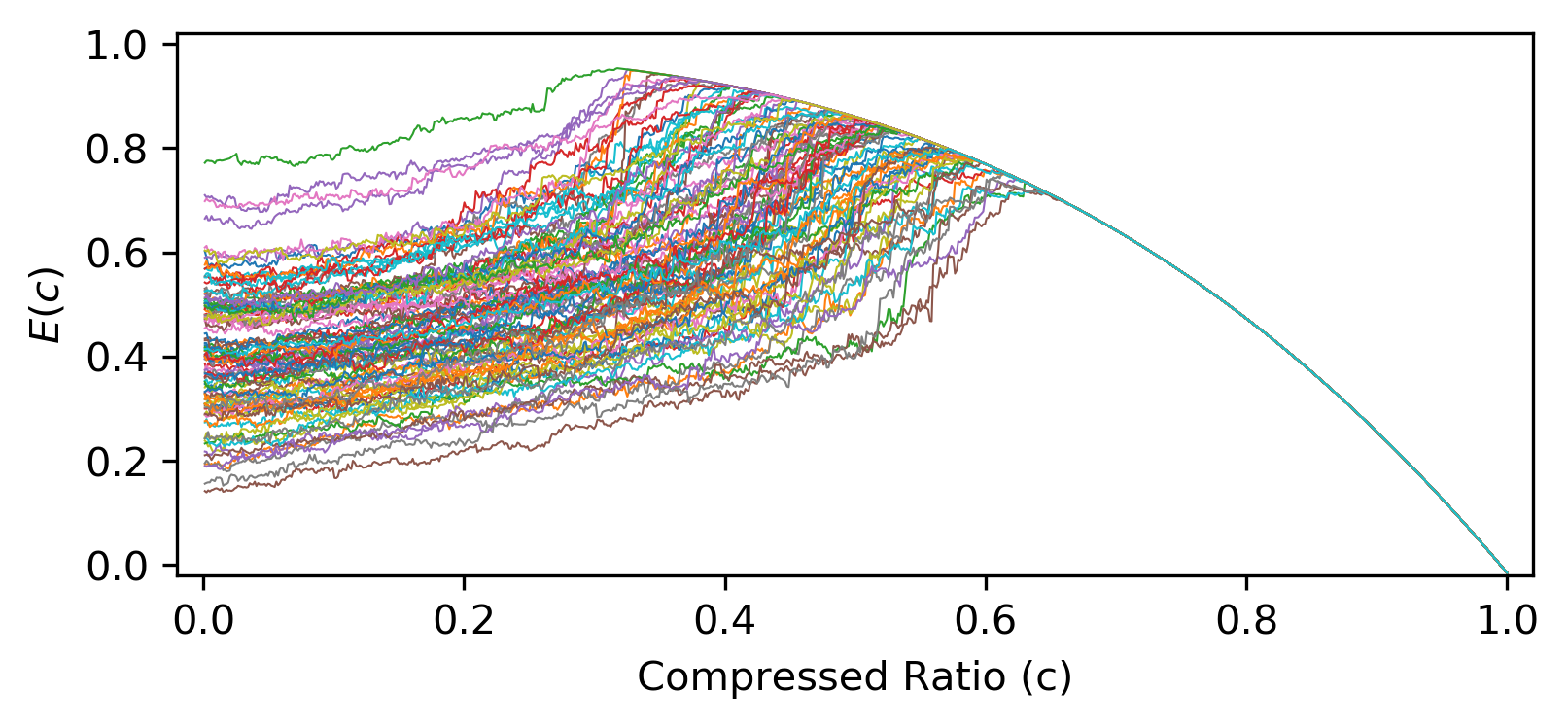}
\subcaption{KMNIST}
\end{center}
\end{minipage}
\caption{Data transmission efficiency of all data in each dataset}
\label{fig:dataset}
\end{figure}

\textbf{Dataset}: 
For the datasets, we used 100 data elements extracted from MNIST \cite{mnist} and Kuzushiji-MNIST (\textit{KMNIST}) \cite{kmnist}.
These data are clearly sparse and meet the requirements of CS.
Figure~\ref{fig:dataset} shows the transmission efficiency $E(c)$ of all data at each compressed ratio in each dataset.
Data taken at random from the dataset were used as input for each step, and the mean $E(c)$ of all data in the dataset was used to evaluate the compression accuracy at each step (To evaluate the performance of RL, the scores for all input states were used to evaluate the model).

\textbf{Environment}: 
To evaluate the utility of the model in sensor nodes, in this experiment, we used Raspberry Pi3\cite{rasp} (Arm Cortex-A53 CPU with a 1.2 GHz clock frequency and 1 GB 450 MHz DDR2 main memory).
In addition, Python 3.9.2 was used for implementation, and linear programming using PuLP\cite{pulp} was adopted for data reconstruction.
Because the accuracy of the model was not dependent on the device, the compression accuracy was evaluated on a MacBook Air (Apple M1, 8Core, 16 GB)\cite{macbookair} to improve the efficiency of the experimental time.

%\begin{comment}
%\end{comment}

\begin{figure}[tb]
\begin{minipage}[t]{0.999\hsize}
%\centering
\begin{center}
\includegraphics[scale=0.62]{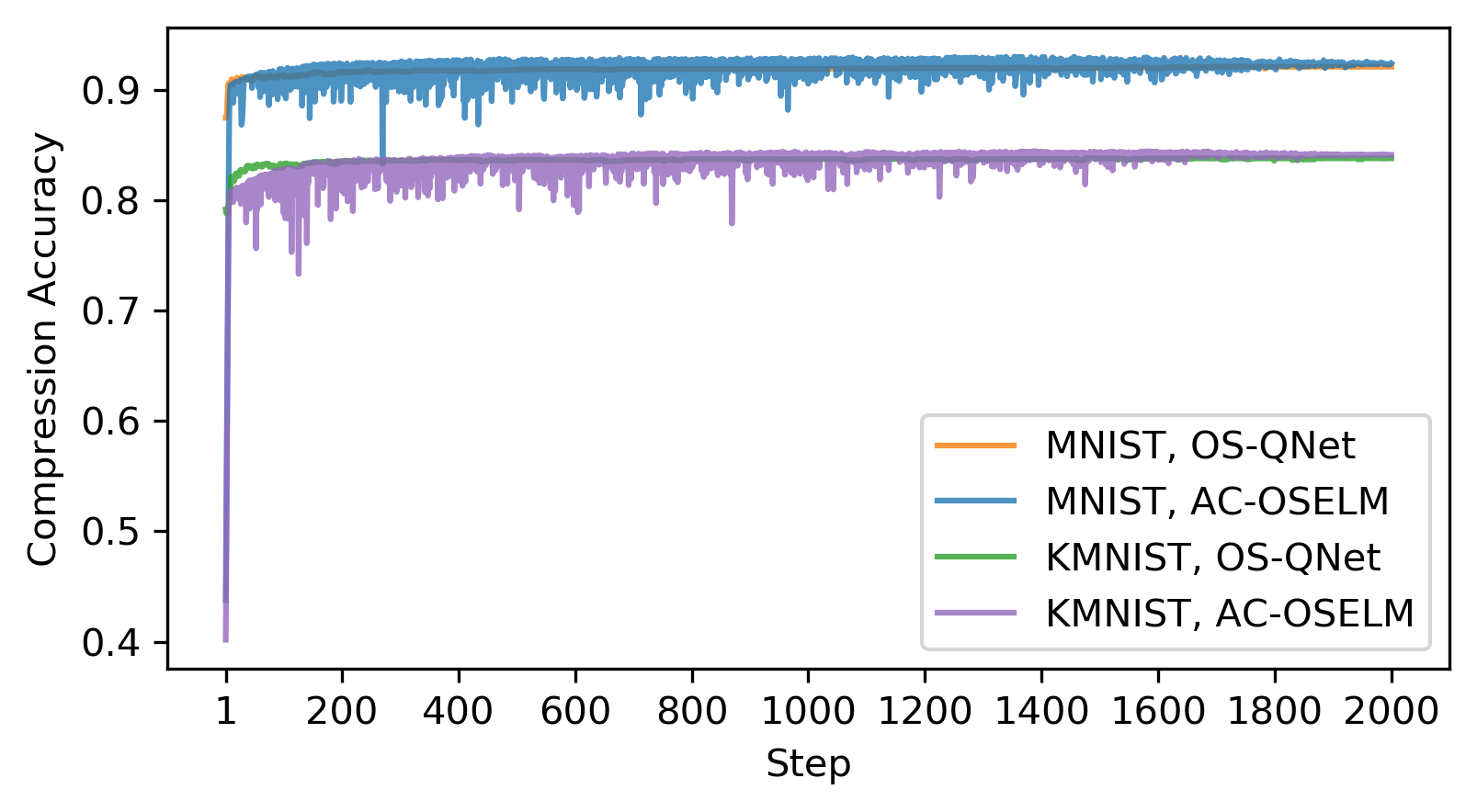}
\end{center}
\end{minipage}
%\vspace{-6mm}
\caption{Compression accuracy at each step when the compression ratio is estimated by each method}
\label{fig:test_score}
%\vspace{-5mm}
\end{figure}

\subsection{Evaluation Result}
\textbf{Compression Accuracy}:
Figure~\ref{fig:test_score} shows the compression accuracy (the mean $E(c)$ of all data in the dataset) at each step when the compression ratio was estimated by each method for each dataset.
In both datasets, the accuracy of each method increased as the steps progressed, indicating that each method learned appropriately.
OS-QNet converged rapidly because its weights were learned analytically.
As the steps progressed, the accuracy of AC-OSELM exceeded that of OS-QNet on each dataset.
This is because AC-OSELM can estimate the compression ratio more finely than OS-QNet.
(The finer the range of compression ratios of OS-QNet, the compression accuracy may become greater, but the computational cost required to estimate the compression ratio increases.
The compression ratio estimation with high computational cost is not suitable for compression at the edge that requires real-time performance.)
Table~\ref{tab:score} lists the maximum compression accuracy and the average compression accuracy from the last to the 10th step for each method on each dataset.
(\textit{Fixed-ratio} shows the maximum accuracy when the compression ratio is held constant for all data.
The accuracy of AC-OSELM/OS-QNet was higher than that of \textit{Fixed-ratio}, indicating that using RL to adaptively estimate the compression ratio for the data is critical for transmission efficiency.)
The maximum/average accuracy in AC-OSELM was up to 0.9/0.3 points higher than that in OS-QNet.
These results indicate that AC-OSELM can achieve the same or better compression ratio estimation performance than OS-QNet.

\begin{table}[b]
%\vspace{-4mm}
\caption{Maximum compression accuracy and average compression accuracy from the last to the 10th step}
%\vspace{-2mm}
\begin{center}
\begin{tabular}{c|c|c|c}
\hline
\hline
%Method~\textbackslash~Dataset& MNIST & KMNIST\\
Dataset~\textbackslash~Method& AC-OSELM & OS-QNet & Fixed-ratio\\
\hline
MNIST&\textbf{0.930} / \textbf{0.923}&0.921 / 0.921&0.908 ($c$=0.432)\\
KMNIST&\textbf{0.844} / \textbf{0.841}&0.838 / 0.838& 0.802 ($c$=0.541)\\
\hline
\end{tabular}
\end{center}
\label{tab:score}
%\vspace{-5mm}
\end{table}

\begin{figure}[t]
\begin{minipage}[t]{0.999\hsize}
\centering
\begin{center}
\includegraphics[scale=0.62]{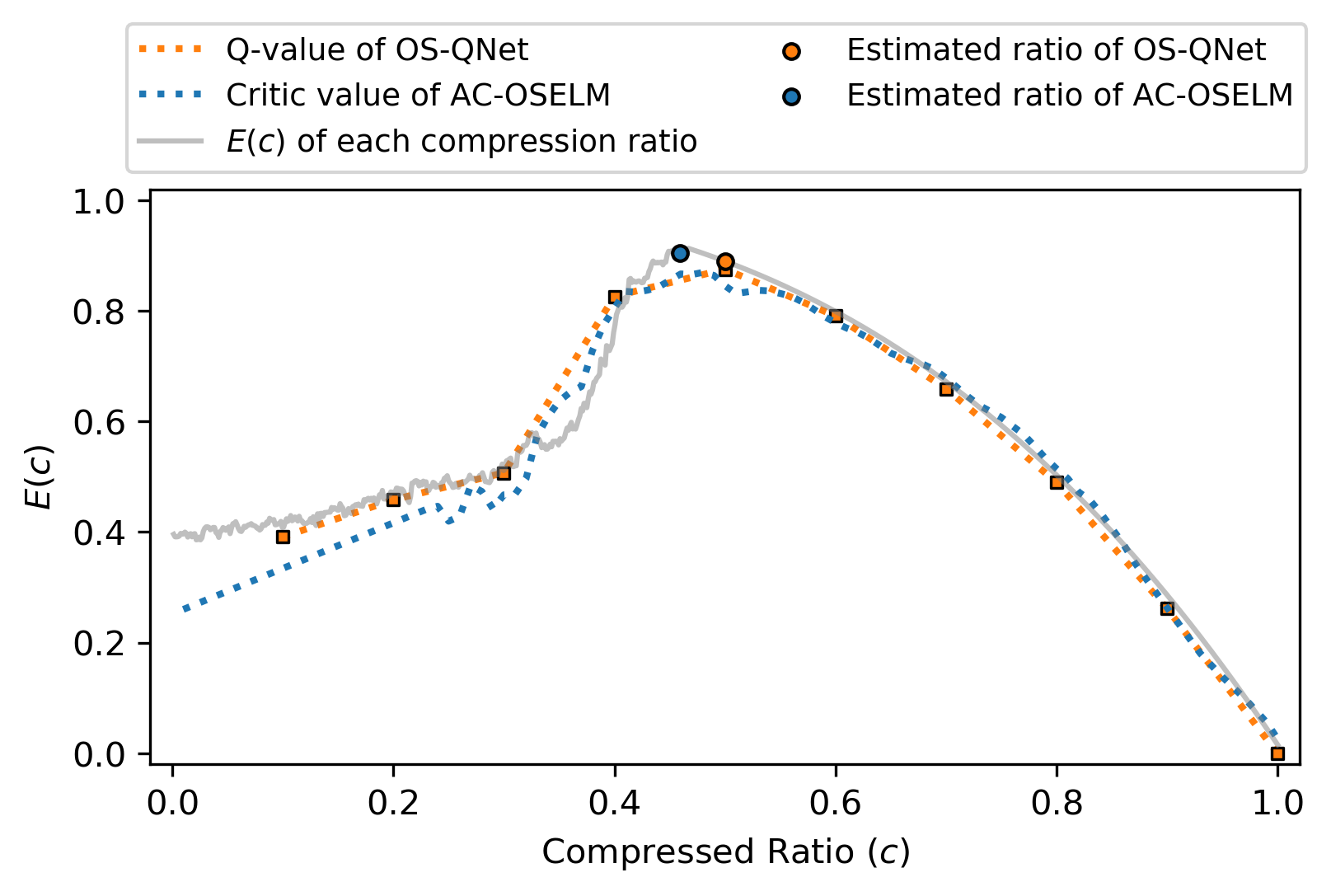}
\end{center}
\end{minipage}
%\vspace{-6mm}
\caption{$E(c)$ and the estimated compressed ratio produced by each model for the one piece of data from MNIST}
\label{fig:ex_estimate}
%\vspace{-5mm}
\end{figure}

\textbf{Estimation Example}:
Figure~\ref{fig:ex_estimate} presents the data transmission efficiency $E(c)$ and the estimated compressed ratio using a trained model of each method for the one piece of data from MNIST (as shown in Figure~\ref{fig:mnist}).
Although the Q-value of OS-QNet and the critic value of AC-OSELM are close (the interval of OS-QNet is rough), AC-OSELM achieves a higher $E(c)$ than OS-QNet. 
This is because AC-OSELM could have a wider selection range of compression ratios and estimate a compression ratio closer to the optimal value.

%Raspi
%\begin{comment}
\begin{table}[b]
%\vspace{-4mm}
\caption{Computational cost of each method}
%\vspace{-2mm}
\begin{center}
\begin{tabular}{c|c|c|c}
\hline
\hline
& Time & \multicolumn{2}{c}{Calculation Time (ms)}\\
\cline{3-4}
Method& Complexity &MNIST&KMNIST\\
\hline
Compression&$\mathcal{O}(D^2)$&1.14&1.38\\
Recovering&$\mathcal{O}(D^3)$&7.35$\times {10}^{4}$&9.79$\times {10}^{4}$\\
\textbf{Inference:}~~~~~~&&&\\
OS-QNet&$\mathcal{O}(|A|(D^2+k^2))$&2.71$\times {10}^{1}$&2.73$\times {10}^{1}$\\
AC-OSELM&$\mathcal{O}(D^2+k^2)$&2.72&2.74\\
\textbf{Learning:}~~~~~~&&&\\
OS-QNet&$\mathcal{O}(n^3+nDk+nk^2)$&6.74$\times {10}^{1}$&6.76$\times {10}^{1}$\\
AC-OSELM&$\mathcal{O}(n^3+nDk+nD^2)$&1.02$\times {10}^{2}$&1.03$\times {10}^{2}$\\
\hline
\end{tabular}
\end{center}
\label{tab:time}
%\vspace{-5mm}
\end{table}
%\end{comment}

\textbf{Computational Cost}:
Table~\ref{tab:time} lists the computational costs associated with the inference and learning for each method.
Here, $|A|$, $D$, $k$, and $n$ indicate the number of action patterns, the number of dimensions of the state, the number of dimensions of the action, and the number of training data pieces, respectively.
Although the computation time of AC-OSELM during learning increased by up to 52.4\% $\left(=\dfrac{1.03\times {10}^{2}}{6.76\times {10}^{1}}-1\right)$ compared to that of OS-QNet, its computation time and complexity during inference were lower than those of OS-QNet.
The computation time required by AC-OSELM to estimate the compression ratio is reduced by up to 90.0\% compared to that required by OS-QNet.
In particular, as $k$ is $1$ in the compression ratio estimation, the computational complexities associated with both data compression in CS and compression ratio estimation in the AC-OSELM are equal ($=\mathcal{O}(D^2)$).
Therefore, estimating the compression ratio using the AC-OSELM does not become a computational bottleneck in the data compression procedure of CS.
By contrast, the computational complexity required to estimate the compression ratio for OS-QNet is greater than that required for data compression.
Considering that data compression on the edge requires real-time performance in inference rather than learning, AC-OSELM-based compression ratio estimation can be considered more suitable for implementation on the edge than the OS-QNet-based counterpart.\par

\section{Limitation and Future Work}
Although OS-QNet/AC-OSELM are small-scale models, they are capable of appropriately estimating compression ratios for less complex data.
However, they may be unable to appropriately learn complex data, for e.g., those that cannot be handled by CS.
In such cases, it is necessary to use a large-scale model such as DRL for the compressed ratio estimation.\par
In learning, recovering data to calculate $E(c)$ is a computational bottleneck, as shown in Table~\ref{tab:time}.
Xu et al. proposed a fast CS recovery method using a generative model based on deep learning\cite{fast_cs_recovery}.
Our future work is to realize these models on a small scale and efficiently approximate $E(c)$ at the edge.

\section{Conclusion}
In this study, we developed AC-OSELM as an efficient RL method for edge devices.
Using AC-OSELM, the selection of actions could be performed faster than with OS-QNet.
We also proposed a data compression method that uses AC-OSELM to estimate the compression ratio in CS.
This allows for rapid compression ratio estimation and process execution on the edge while achieving adaptive compression for the acquired data. 
To evaluate the compression ratio estimation performance of the proposed AC-OSELM, it was compared with OS-QNet.
The experimental results demonstrated that the proposed method outperformed the compared method with regard to the compression performance and computational cost of estimating the compression ratio.
This confirms that the AC-OSELM-based compression-ratio estimation method is suitable for implementation on edge devices.

\bibliographystyle{IEEEtran.bst}
\input{main.bbl}

\end{document}

%% file: main.bbl
% Generated by IEEEtran.bst, version: 1.12 (2007/01/11)